# Building Entity Association Mining Framework for Knowledge Discovery

Anshika Rawal
*Fidelity Investments*
Bengaluru, India
anshika.rawal@fmr.com

Abhijeet Kumar
*Fidelity Investments*
Bengaluru, India
abhijeet.kumar@fmr.com

Mridul Mishra
*Fidelity Investments*
Bengaluru, India
mridul.mishra@fmr.com

*Abstract*— Extracting useful signals or pattern to support important business decisions for example analysing investment product traction and discovering customer preference, risk monitoring etc from unstructured text is a challenging task. Today, overwhelming amount of written communication data such as legal filings, or business and financial news data is consumable. Capturing interaction of entities or concepts and association mining is a crucial component in text mining, enabling information extraction and reasoning over and knowledge discovery from text. Furthermore, it can be used to enrich or filter knowledge graphs to guide exploration processes. Knowledge graphs are useful tool which can be leveraged for such descriptive analytics and uncover hidden stories in the text.

In this paper, we introduce a domain independent pipeline i.e., generalised framework to enable document filtering, entity extraction using various sources (or techniques) as plug-ins and association mining to build any text mining business use-case and quantitatively define a scoring metric for ranking purpose. The same can be utilized for graph generation of extracted entities to discover association trends of entities or concepts of interest and clusters of themes in the built entity cooccurrence network. Further, the proposed solution can be leveraged in different domain specific use-cases with minimal configurations.

The proposed framework has three major components a) **Document filtering:** filtering documents/text of interest (lexical search or semantic search) from massive amount of texts b) **Configurable entity extraction pipeline:** to accommodate different domain independent use-cases. Currently, this module supports multiple entity extraction techniques i.e., i) DBpedia Spotlight, ii) Spacy NER (Named Entity Recognition), iii) Custom Entity Matcher, iv) Phrase extraction (or dictionary) based c) **Association Relationship Mining:** To generates co-occurrence graph to analyse potential relationships among entities, concepts. Further, co-occurrence count based frequency statistics (pair count at sentence, paragraph, or article level) provide a holistic window to observe association trends or buzz rate in specific business context. The paper demonstrates the usage of framework as fundamental building box in two financial use-cases namely *brand product discovery* and *vendor risk monitoring*.

We aim that such framework will remove duplicated effort, minimize the development effort, and encourage reusability and rapid prototyping in association mining business applications for institutions.

## I. INTRODUCTION

Analysts in finance domain are faced with the challenging task of analysing very high volume of unstructured data in order to make critical decision related to risk compliance, analysing investment product traction and customer experience etc. Since most of the data records in unstructured and narrative form, the quality and availability of natural language processing tools has become critical to perform high-level tasks such as information extraction, and knowledge discovery.

Currently, Significant amount of time is spent on gathering and understanding this data before knowing whether the information is useful or not. While these documents are increasing massively on day-to-day basis, it has become almost impossible for manual processes to keep up which has spurred the need to automate these analysis processes.

When tasked with extracting insights from huge amounts of financial data, automatic entity recognition is an important first step, and then finding association relationship on top takes your analysis to the next level.

Named Entity Recognition (NER) is a core component in Information Retrieval which detects the occurrence of named entities in unstructured text and classify each extracted entity into a particular category. The "Named Entity" term was introduced in 1996 at the 6th Message Understanding Conference (MUC) when the information extraction from unstructured text became a critical task. In general, there are two approaches: knowledge engineering and machine learning methods. Knowledge engineering approaches utilized expert knowledge to recognize named entity in the text. This method is usually rule-based. In machine learning techniques, we train an algorithm to identify named entities.

The present paper utilizes more than one NER methodologies such as Spacy NER, Custom Entity Matcher etc. to improve the performance as well as enrich the extracted entity set by using Linked Open Data cloud through DBpedia [1]. DBpedia Spotlight allows users to configure the annotations to their specific needs through the DBpedia Ontology and quality measures such as prominence, topical pertinence, contextual ambiguity and disambiguation confidence. It distinguishes itself from the other annotators, which are either limited to one or two languages, or only available as paid services. Where Dbpedia[1] is large-scale, Multilingual Knowledge base extracted from Wikipedia which supports the annotation of 3.5 million entities and concepts from more than 320 classes.

The proposed Framework is focused on automatically identifying named entities (NER) as first step in the pipeline followed by detecting the association between entities extracted from legal filings, or business and financial news data. In general, the goal is to reduce the amount of information to provide humans an overview and enable the generation of new insights. One such representation are knowledge graphs. Traditionally, detecting association relationship between entities involves adopting co-occurrence methods. The approach used here is to construct a graph where the entities in the document are nodes and link between them represents if there is any association between the entities, The underlying assumption behind two entities association is that those two entities must co-occur within a certain window (a set of words occurring around a given entity). Further, frequencies of co-occurrences are calculated, and the network is analysed to find association trends, discover important entities or concepts and clusters of themes in the network.

This paper is organized as follows. In Section 2, we describe the background of this paper. In Section 3, we present our method for mining relationship associations. We also describe the use case built on top of the knowledge graph and its meta data. Typical use cases assume some prior knowledge of the data and support the user by retrieving and arranging the relevant extracted information. In Section 4, we present results of experiments to mine relationship associations from a set of semantic graphs representing different use-cases. We conclude this paper with a discussion of related work and a summary.

## II. BACKGROUND LITERATURE

Entity extraction has been the core subtask when it comes to information extraction, especially now when technologies allow the storage of and processing of massive data. Traditionally, research in this area is founded in computational linguistics, where the goal is to parse and describe the natural language with statistical rule-based methods. The foundation for that is to correctly tokenize the unstructured text, assign part-of-speech tags, and create a parse tree that describes the sentence's dependencies and overall structure. Using this information, linguists defined rules that describe typical patterns for named entities. Handcrafted rules were soon replaced by machine learning approaches that use tags mentioned above and so-called surface features. These surface features describe syntactic characteristics, such as the number of characters, capitalization, and other derived information.

To extract information related to extremely heterogeneous structured company names from news articles, loster et al. [2] has presented a dictionary- based approach in "Improving Company Recognition from Unstructured Text by using Dictionaries". They mentioned the challenges of their proposed system in context of when applied in a large financial organization [3].

Apart from the targeted NER methods for specific task, there are open-source libraries which enables users to quickly annotate their data without having built the specific NER module. The following overview depicts the few most successful open-source libraries. The Natural Language Toolkit (NLTK) first released in 2001, is one of the most popular Python libraries for natural language processing [4]. It provides a wealth of easy-to-use API for all traditional text processing tasks and named entity recognition capabilities. The Apache OpenNLP project is a Java library for the most common text processing tasks and was first released in 2004 [5]. It provides implementations of a wide selection of machine learning-based NLP research designed to extend data pipelines built with Apache Flink or Spark. The Stanford NLP research group released their first version of CoreNLP in 2006 as a Java library with Python bindings [6]. It is actively developed and provides NLP tools ranging from traditional rule-based approaches to models from recently published state-of-the-art deep learning research. Only recently in 2015, spaCy was released as a Python/Cython library, focusing on providing high performance in terms of processing speed. It also includes language models for over 50 languages and a growing community of extensions. After an initial focus on processing speed, the library now includes high-quality language models based on recent advances in deep learning.

As indicated by prior studies [8], incorporating other knowledge sources into NER approaches can produce better results for results for ranking the results for relationships between two entities as any standalone tool or method of NER has its own limitations in terms of disambiguation and heterogeneous structure of entities and same entity having different representations in terms of acronyms and abbreviations. Several of those publicly available and covers a significantly high number of linked entities and concepts are Wikipedia Miner [9], YAGO [7], Tagme [10], AIDA based on Stanford DBpedia [11].

In addition to entity extraction, association mining of these extracted entities or concepts provides the hidden insights. Simple co-occurrence-based approaches are one of few methods to establish a relationship or association among these entities. Borislav Popov et al. [12] has presented a framework in "Co-occurrence and Ranking of Entities" called CORE in addition of their existing system called KIM [13]. In this paper, they have used predefined set of relations to get the cooccurrence of entities which is unlikely to cover all the relations mentioned in real world dataset. In addition to simple co-occurrence-based approaches to measuring the relationship between entities, rule-based methods using syntactic patterns [14,15] The presented paper is motivated by the concept of utilizing knowledge sources for enriching entity set which is intended to result in dense co-occurrence graph along with association relationships.

## III. PROPOSED FRAMEWORK

The proposed entity association framework brings out the knowledge by reducing feature space with focused on named entities recognition and their co-occurrence accompanied by other metadata. This framework supports the full text search to track any information in document filtering module similar to traditional information retrieval systems which enables user to retrieve the relevant documents from large noisy corpus using a user defined query.

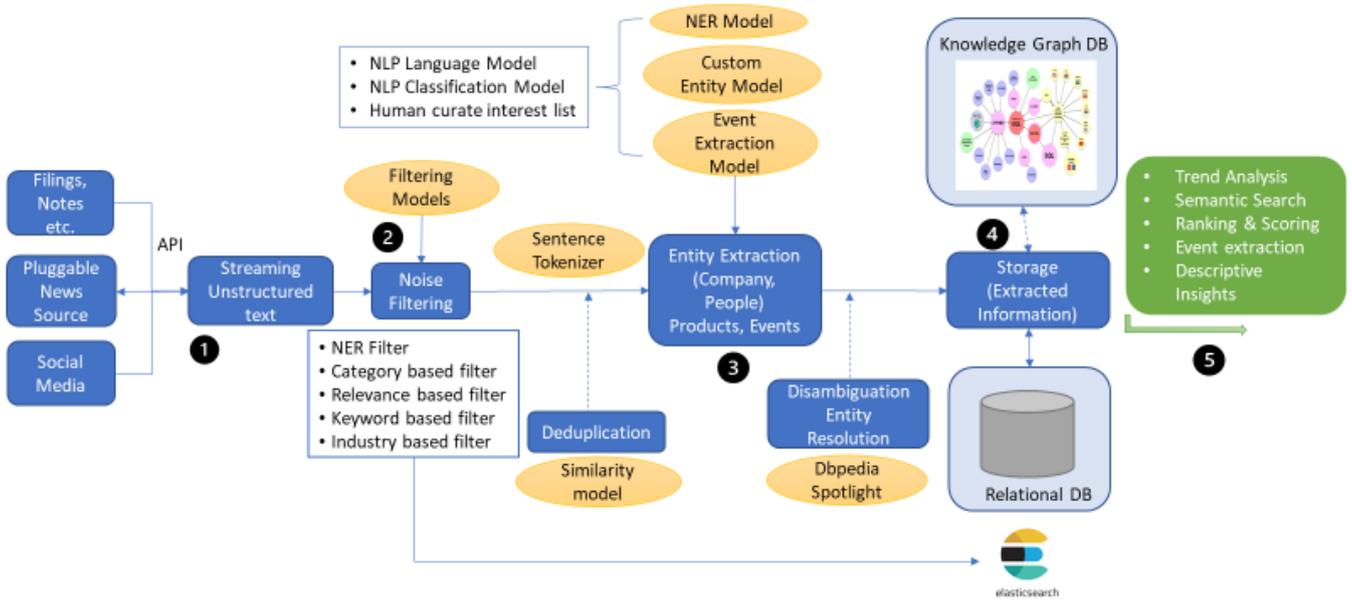

Fig 1. Architecture of proposed framework

In this framework one can query about co-occurred entities with given an entity, which can further be restricted by entity type or timeline of their cooccurrence. This module enables user to track correlation between different type of entities along with their trend deviation based on their cooccurrence. The proposed framework has several components as shown in fig. 1 which are penned below.

*A. Document filtering:*

This module is designed to filter out the noisy data and extract the relevant corpus based on user defined query. Currently, this module has two strategies one is lexical match-based filter another is semantic relevance-based filter. Lexical match-based filter utilizes the spacy's PhraseMatcher for matching large terminology lists and stores the doc object of the articles in Docbin format efficiently. This Lexical match-based filter generates the matched keyword from user query along with their count. Another approach, semantic relevance-based filter uses the pretrained sentence-transformer model[16] to generate the embedding of articles as well as the user query. In this approach, cosine score depicts the relevance of a particular article to the user query.

*B. Deduplication:*

The case where different media houses publish the same news occurs more often with increasing data and news source, resulted duplicated information introduces the bias when it comes to downstream analysis. This module provides the clustering capability which ultimately enables user to get rid of duplicated news. Additionally, document clustering also facilitates to discover groups of similar documents/news in a corpus for meaningful segregation. We have used Hierarchical clustering used to group news articles in clusters based on their embedding similarity. Currently embedding are being generated by two methods (i) Doc2vec embeddings which is trained on news corpus. (ii) Sentence transformer's embedding which are obtained by using pretrained language models. This module also supports finetuning clusters by Silhouette score for range of thresholds.

*C. Entity Extraction:*

This is another configurable module in Framework which is intended to extract entities and concepts along with user defined keywords. it supports multiple prebuilt NER techniques, which can be used either as plugins or in parallel to enhance the performance as entity extraction is a fundamental block to derive insights in any specific use-case. Currently it supports three types of entity extraction technique.

DBpedia spotlight [17]: It is a tool which enables the user to interlink and perform semantic query among high quality knowledge bases to extract information from unstructured data. It performs four steps i.e., i) Spotting ii) Candidate Selection iii) Disambiguation iv) Filtering. It annotates the document with DBpedia URIs which helps to resolve the named entity disambiguation problem. Here is an example of ambiguity problem, 'SEC' and 'Stock Exchange Commission' are two different representations of same entity. In this case DBpedia spotlight links both the entities with same DBpedia URI.

Spacy NER [18]: Spacy is one of the popular library for NER which has language model for more than 50 languages. This prebuilt technique can recognize various types of named entities in a document such as name, location, organization etc. As the performance depends on the example spacy models are trained on so it might require fine tuning according the specific use-case .

Custom Entity Matcher: It is rule-based spacy Matcher which uses POS tags to identify the specific patterns such as investment product names mentioned in the documents. The generic pattern this module uses is [{'POS': {"IN": ["NOUN", "PROPN", "ADJ", "VERB", "ADV"]}, 'OP': '+', 'IS_STOP': False}, {"LEMMA": {"IN": value}}] where value is list of keywords corresponding to each investment product category. Currently this module is implemented for four types of investment products including Fund, Bond, ETF, Derivatives and easily configurable to add more investment products or entity of interest.

Financial Event Detection: This module enriches the extracted meta data by extracting the financial event in a document using AWS offering (Comprehend Event Extraction Service). This service detects following events "BANKRUPTCY", "EMPLOYMENT", "CORPORATE_ACQUISITION", "INVESTMENT_GENERAL", "CORPORATE_MERGER", "IPO". For example, in snippet of a news "A spokesperson for Schwab, which acquired TD Ameritrade in October 2020, confirmed the departures", this module will extract "ACQUISITION" as the financial event.

*D. Entity Association:*

The **entity association** module enables user to query about co-occurring entities and the trend of their occurrence within a timeline or restricted subset of the corpus by other features like entity type. This new feature space of documents focused on named entities is more interpretable and insightful than traditional information retrieval system which process full text to gather any information. It also allows user to effectively track the relationship between entities and time point and other metadata to filter the further results.

The extracted knowledge is combination of semantic as well as of statistical information. Where semantic information details about extracted entities along with their classification, The co-occurrence is entirely statistical information although it has hidden semantic meaning. This enables user to search and discover useful new insights related to different domain specific use-cases with minimal configurations.

## IV. EXPERIMENTS AND RESULTS

There were two financial use-cases where the framework was employed to discover knowledge and insightful trends.

1) Brand Product Discovery
2) Vendor Risk Monitoring.

The development of both use-cases took the minimal effort as all the complexity was leveraged from underlying modules of framework. This section briefly describes both use-cases along with experiments, results, and analysis of co-occurrence graphs.

*A. Dataset*

There are two different datasets gathered keeping in mind the different use-cases. First, we collected ~30K investment news published over last 5 years from Ignite [19] and FundFire [20] websites by web scrapping. Both are media sources who publishes news related to investment products in financial industry. Secondly, we collected a dataset of 70K financial news article from a news aggregator Aylien [21] by filtering trusted media sources only.

*B. Usecase: Brand Product Discovery*

We extracted investment product names and associated brands from the investment product news corpus to perform competitor analysis and analyze emerging trend for any specific investment theme. we generate a cooccurrence graph of investment firms and their trending or launched product associated over the timeline and traction about it in investment news feed.

| | 2016 | 2017 | 2018 | 2019 | 2020 | 2021 | 2022 | |
|---|---|---|---|---|---|---|---|---|
| Morningstar | 21 | 13 | 21 | 17 | 9 | 37 | 49 | 167 |
| Fidelity | 12 | 19 | 14 | 15 | 19 | 22 | 14 | 115 |
| BlackRock | 13 | 12 | 20 | 9 | 21 | 15 | 15 | 105 |
| Vanguard | 8 | 20 | 15 | 16 | 14 | 7 | 15 | 95 |
| Bloomberg | 4 | 21 | 15 | 4 | 8 | 14 | 17 | 83 |
| Invesco | 6 | 7 | 19 | 7 | 15 | 13 | 5 | 72 |
| Blackstone | 0 | 14 | 10 | 3 | 10 | 7 | 6 | 50 |
| Goldman | 6 | 8 | 4 | 5 | 12 | 4 | 10 | 49 |
| JP Morgan | 7 | 7 | 8 | 9 | 8 | 2 | 6 | 47 |
| SSGA | 8 | 4 | 7 | 8 | 6 | 6 | 5 | 44 |
| T Rowe | 9 | 2 | 2 | 7 | 10 | 10 | 4 | 44 |
| FT | 3 | 7 | 3 | 4 | 15 | 1 | 10 | 43 |
| Pimco | 9 | 10 | 2 | 2 | 5 | 5 | 5 | 38 |
| American Century | 1 | 1 | 4 | 11 | 10 | 6 | 2 | 35 |
| KKR | 0 | 4 | 5 | 2 | 15 | 6 | 0 | 32 |
| WisdomTree | 10 | 3 | 2 | 1 | 5 | 9 | 2 | 32 |

Fig 2. Example: Snippet of Brand Product Discovery

Fig 2. ranks all the investment brands with year-on-year count occurrences of trending, popular or launched products at sentence level. The system developed consists of 4 major steps: At first, all the news articles with trending or popular products were filtered. Next, it segments article into sentences and further selects specific sentences with the target language. As step 3, system runs entity extractor and extract all companies and investment products names using spacy matcher. Finally, association miner links a brand to all co-occurring investment products. Fig. 3 shows investment product co-occurring with Fidelity investment in news with launch or trending language (lexical match).

| Product_name | 2016 | 2017 | 2018 | 2019 | 2020 | 2021 | 2022 | sentences Total |
|---|---|---|---|---|---|---|---|---|
| nontransparent etf | 0 | 0 | 0 | 1 | 6 | 2 | 1 | 10 |
| bitcoin etf | 0 | 0 | 0 | 0 | 0 | 3 | 1 | 4 |
| portfolio shielding etf | 0 | 0 | 0 | 0 | 3 | 1 | 0 | 4 |
| magellan fund | 0 | 0 | 0 | 0 | 3 | 0 | 0 | 3 |
| target date fund | 0 | 0 | 0 | 2 | 0 | 0 | 1 | 3 |
| bitcoin fund | 0 | 0 | 0 | 0 | 1 | 0 | 1 | 2 |
| hartford fund | 0 | 1 | 0 | 0 | 0 | 1 | 0 | 2 |
| spartan fund | 2 | 0 | 0 | 0 | 0 | 0 | 0 | 2 |
| spot bitcoin etf | 0 | 0 | 0 | 0 | 0 | 0 | 2 | 2 |
| bitcoin investment fund | 0 | 0 | 0 | 0 | 0 | 2 | 0 | 2 |
| semitransparent etf | 0 | 0 | 0 | 0 | 2 | 0 | 0 | 2 |
| esg fund | 0 | 0 | 0 | 2 | 0 | 0 | 0 | 2 |
| smart beta etf | 1 | 0 | 1 | 0 | 0 | 0 | 0 | 2 |

Fig 3. Fidelity: Product co-occurrence frequency (in news)

A snippet of co-occurrence graph is shown in Fig. 4 where Invesco, Vanguard, and Fidelity (green nodes) connected to variety of investment products (orange nodes) with count of co-occurrence and matched keywords as edge properties. This allows user to either discover the competitors connected to certain type of investment theme or vice versa. For example, Fidelity has been mentioned with *bitcoin etf, bitcoin fund, bitcoin investment fund, and spot bitcoin etfs* which are products in cryptocurrency space. Similarly, one can explore news around ESG funds launched by Fidelity in 2019 along with linked competitors.

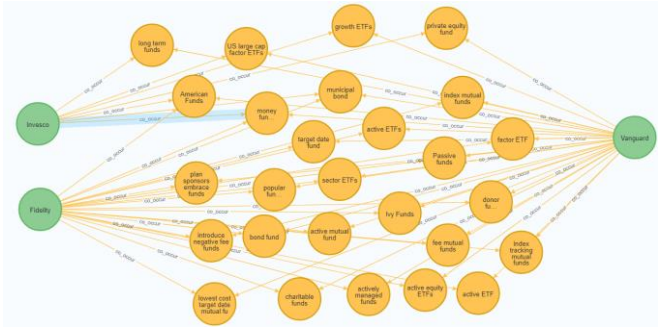

Fig 4. Co-occurrence Graph: Brand Product Association

### C. Usecase: Vendor Risk Monitoring

A potential application was explored in building vendor risk monitoring system where the proposed framework was employed to generate insight about risk associated with a given vendor in terms of any legal, compliance or governance risk over time. For the experiments, we defined risk entities of 100 unique risk diction. Employing the proposed framework, document filtering component filters news matching risk diction as a first step. Further, entity extraction component breaks text into sentences and extracts all *ORG* entities using spacy tool and converts risk diction as risk entities at sentence level. In next steps, entity association component establishes the association linking and create co-occurrence graphs. Such systems may facilitate vendor managers to generate vendor risk profile comprehensively as well as rank them based on risk associated with vendors as generated in Fig. 6.

|  | 2016 | 2017 | 2018 | 2019 | 2020 | 2021 | 2022 | Total |
|---|---|---|---|---|---|---|---|---|
| Morgan stanley | 14 | 88 | 86 | 36 | 78 | 29 | 35 | 366 |
| Reuters | 3 | 95 | 101 | 40 | 45 | 47 | 33 | 364 |
| Wells Fargo | 15 | 109 | 59 | 48 | 51 | 21 | 9 | 312 |
| Blackrock | 9 | 19 | 33 | 28 | 26 | 71 | 13 | 199 |
| JP Morgan | 5 | 64 | 43 | 14 | 15 | 29 | 11 | 181 |
| Merrill Lynch | 7 | 47 | 50 | 23 | 10 | 6 | 4 | 147 |
| BNY Mellon | 10 | 14 | 5 | 21 | 25 | 20 | 19 | 114 |
| Morningstar | 1 | 13 | 7 | 27 | 20 | 28 | 18 | 114 |
| Bloomberg | 2 | 16 | 18 | 18 | 22 | 13 | 7 | 96 |
| State Street | 1 | 10 | 3 | 9 | 12 | 9 | 0 | 44 |

Fig 5. Example: Snippet of Vendors Risk Ranking

The above heatmap depicts the count statistics of associated diction and depends on number of articles and buzz around certain risk event in media sources. Further to assist, risk analysts or vendor managers can analyze the specific news which accounted to counts in each cell of heatmap. A different view of risk diction association for a specific company *Morgan Stanley* can also be shown (Fig. 7).

|  | 2016 | 2017 | 2018 | 2019 | 2020 | 2021 | 2022 | Total |
|---|---|---|---|---|---|---|---|---|
| lawsuit | 3 | 13 | 10 | 3 | 19 | 4 | 6 | 58 |
| litigation | 2 | 11 | 5 | 1 | 4 | 0 | 1 | 24 |
| breach | 2 | 1 | 1 | 1 | 8 | 4 | 5 | 22 |
| harassment | 0 | 6 | 7 | 5 | 3 | 0 | 0 | 21 |
| allegations | 0 | 2 | 12 | 2 | 4 | 1 | 0 | 21 |

Fig 6. Top 5 Risk Diction Association: Morgan Stanley (Frequency)

Like previous use-case, a snippet of co-occurrence graph is shown below (Fig. 8) which consists of risk nodes and vendor nodes and edges quantified by number of co-occurrences. One can observe all the companies (vendors) linked to various risks like *layoffs, data breach, litigation (lawsuits)*.

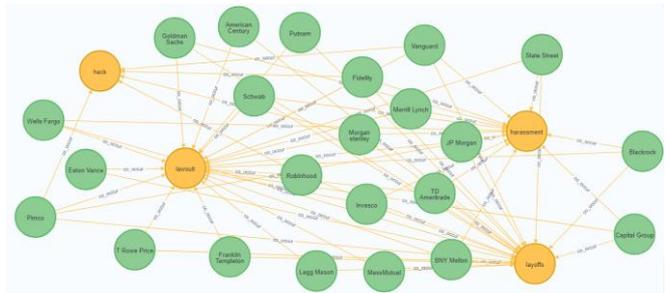

Fig 7. Co-occurrence Graph: Vendors-Risk graph

Such co-occurrence graph with variety of entities can be exploited for many downstream tasks like representation learning, node similarity and clustering. The process flow of generic entity interaction graph (massive graph with interaction of entities) utilizing the developed framework is shown below.

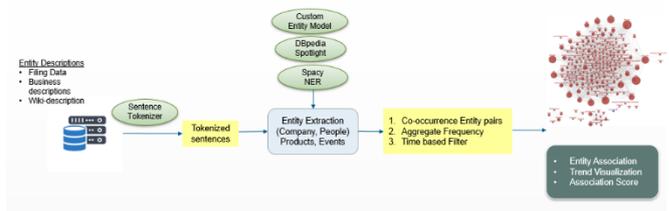

Fig 8. Entity Association Mining Process Flow

## V. CONCLUSION AND FUTURE WORK

In this paper, we attempted to implement a generic entity association mining framework to resolve the issue of duplicated work and loss of resource for multitude of NLP tasks in information mining and discovery space. This paper presented a configurable pipeline comprised of document filtering, entity extraction, entity association mining by obtaining cooccurrence of extracted entities in news article text data. The paper provides an overview of different pluggable techniques used in various parts of pipeline. We emphasized on the applications of the proposed framework using two different use-cases of finance namely brand product discovery and vendor risk monitoring. Also, the paper describes construction of the generated knowledge graph and heat maps that can be visualized and explored to support decisions, investigation, and analysis of entities in respective use-cases. Our proposed system is built to work on historical data dump which can be scaled in an efficient way where the integration of continuously flowing new data can be accommodated seamlessly to update the knowledge graphs and trends.

Currently, significance of an entity pair is being measured in simplistic way, frequency of cooccurred entities is counted directly without any normalization with respect to the size of news articles. More sophisticated methodology can be opted in future work to rank entities and timeline trends can be adjusted accordingly. Additionally, entity disambiguation is being performed only by DBpedia which can be extended to other entities recognition plugin modules by building a domain specific ontology. Also, we are looking forward to employing the framework as a tool to develop insights and trends on any textual corpus and further finetune it for various business use-cases.